\documentclass{MITcsail}

\usepackage{amsmath,amsfonts,bm}

\def\eqref#1{equation~\ref{#1}}

\def\1{\bm{1}}

\DeclareMathAlphabet{\mathsfit}{\encodingdefault}{\sfdefault}{m}{sl}
\SetMathAlphabet{\mathsfit}{bold}{\encodingdefault}{\sfdefault}{bx}{n}

\newcommand{\KL}{D_{\mathrm{KL}}}

\usepackage{hyperref}
\usepackage{url}

\usepackage{amssymb}
\usepackage[english]{babel} 
\usepackage{booktabs}
\usepackage{multicol}
\usepackage{multirow}
\usepackage{graphicx}
\usepackage{tabularx}
\usepackage{amsthm}
\usepackage{subcaption}
\usepackage[most]{tcolorbox}
\usepackage{xcolor}

\definecolor{qwen3-4B}{RGB}{31,119,180}
\definecolor{qwen2.5-3B}{RGB}{174,199,232}
\definecolor{qwen3-1.7B}{RGB}{140,86,75}

\title{No Free Lunch: Rethinking Internal Feedback for LLM Reasoning}

\author
{Yanzhi Zhang~$^{1,2,3}$\footnote{Equal Contribution.}, Zhaoxi Zhang~$^{4,5}$\footnotemark[1], Haoxiang Guan~$^{1}$, Yilin Cheng~$^{1}$, Yitong Duan~$^{1,5}$,\\
\textbf{Chen Wang~$^{1}$}, \textbf{Yue Wang~$^{1,5}$}, \textbf{Shuxin Zheng~$^{1,5}$}, \textbf{Jiyan He~$^{1,5}$\footnote{Correspondence. E-mail: \texttt{v-hjy@zgci.ac.cn}}}\\
\vspace{1em}
\normalfont{\small $^{1}$Zhongguancun Academy}\\
\normalfont{\small $^{2}$Academy of Mathematics and Systems Science, Chinese Academy of Sciences}\\
\normalfont{\small $^{3}$University of Chinese Academy of Sciences}\\
\normalfont{\small $^{4}$Peking University}\\
\normalfont{\small $^{5}$Zhongguancun Institute of Artificial Intelligence}\vspace{2em}
}

\begin{document}
\maketitle
\thispagestyle{firstpagestyle} 

\begin{abstract}
Reinforcement learning has emerged as a powerful paradigm for post-training large language models (LLMs) to improve reasoning. Approaches like Reinforcement Learning from Human Feedback (RLHF) and Reinforcement Learning with Verifiable Rewards (RLVR) have shown strong results, but they require extensive external supervision. We investigate an alternative class of methods, Reinforcement Learning from Internal Feedback (RLIF), which relies solely on intrinsic model-derived signals instead of external rewards. In particular, we leverage unsupervised reward proxies such as token-level entropy, trajectory-level entropy, and self-certainty. Our theoretical analysis shows these internal objectives are partially equivalent, and we empirically evaluate various RLIF strategies on challenging math reasoning benchmarks. Experimental results demonstrate that RLIF can boost the reasoning performance of base LLMs at the beginning phase of the training, matching or surpassing RLVR techniques on these tasks. However, when training progresses, performance degrades even below the model before training. Moreover, we find that RLIF yields little improvement for instruction-tuned models, indicating diminishing returns of intrinsic feedback once an LLM is already instruction-tuned. We further analyze this limitation by mixing model weights and explain the reason of RLIF's training behaviors, providing practical guidelines for integrating internal feedback signals into LLM training. We hope our analysis of internal feedback will inform more principled and effective strategies for LLM post-training.
\end{abstract}

\section{Introduction}
Recent advances in large language models (LLMs) \citep{glm2024chatglm,guo2025deepseek, liu2024deepseek, touvron2023llama} have been driven not only by architectural innovations and scaling strategies \citep{chen2024internvl}, but also by increasingly sophisticated reinforcement learning (RL) techniques. While pre-training has delivered remarkable progress, its effectiveness is approaching a plateau, necessitating alternative methods for further improvement. One widely adopted approach has been Reinforcement Learning from Human Feedback (RLHF) \citep{bai2022training} to align model outputs with human preferences. In parallel, methods that leverage automatically verifiable signals—referred to as Reinforcement Learning with Verifiable Rewards (RLVR) \citep{yu2025dapo,hu2025reinforce++,shao2024deepseekmath}—have been developed to further enhance reasoning capabilities in domains such as mathematics and code. For instance, following the release of DeepSeek-R1 \citep{guo2025deepseek} (which employed the GRPO algorithm~\citep{shao2024deepseekmath}) and more recently Qwen3~\citep{yang2025qwen3}, RLVR has demonstrated promising improvements by directly optimizing from pretrained models and bypassing traditional supervised fine-tuning (SFT).

Despite these advances, \citet{yue2025doesreinforcementlearningreally} demonstrates that the RLVR approach struggles to surpass the inherent capability limits of large models: in domains such as mathematics and code reasoning, the pass@N metrics show no improvement; only pass@1 has increased, this conclusion can also be found in \citep{shao2024deepseekmath}. The phenomenon suggests that RLVR primarily shifts the model's output distribution, leading to more efficient sampling while simultaneously narrowing the scope of its reasoning boundaries. Recent works~\citep{2025selfcertainty,zhihu2025entropy} have introduced the concept of Reinforcement Learning with Internal Feedback (RLIF), that is, reinforcement learning without an externally given reward signal. RLIF methods take a more direct approach by optimizing with a reward signal derived directly from a scalar value related to the model's output distribution, rather than relying on external signals as rewards. The key advantage of this unsupervised reinforcement learning approach is its lack of reliance on labeled ground truth data, thereby reducing data requirements and facilitating training.

\begin{figure}[t]
    \centering
    \includegraphics[width=\linewidth]{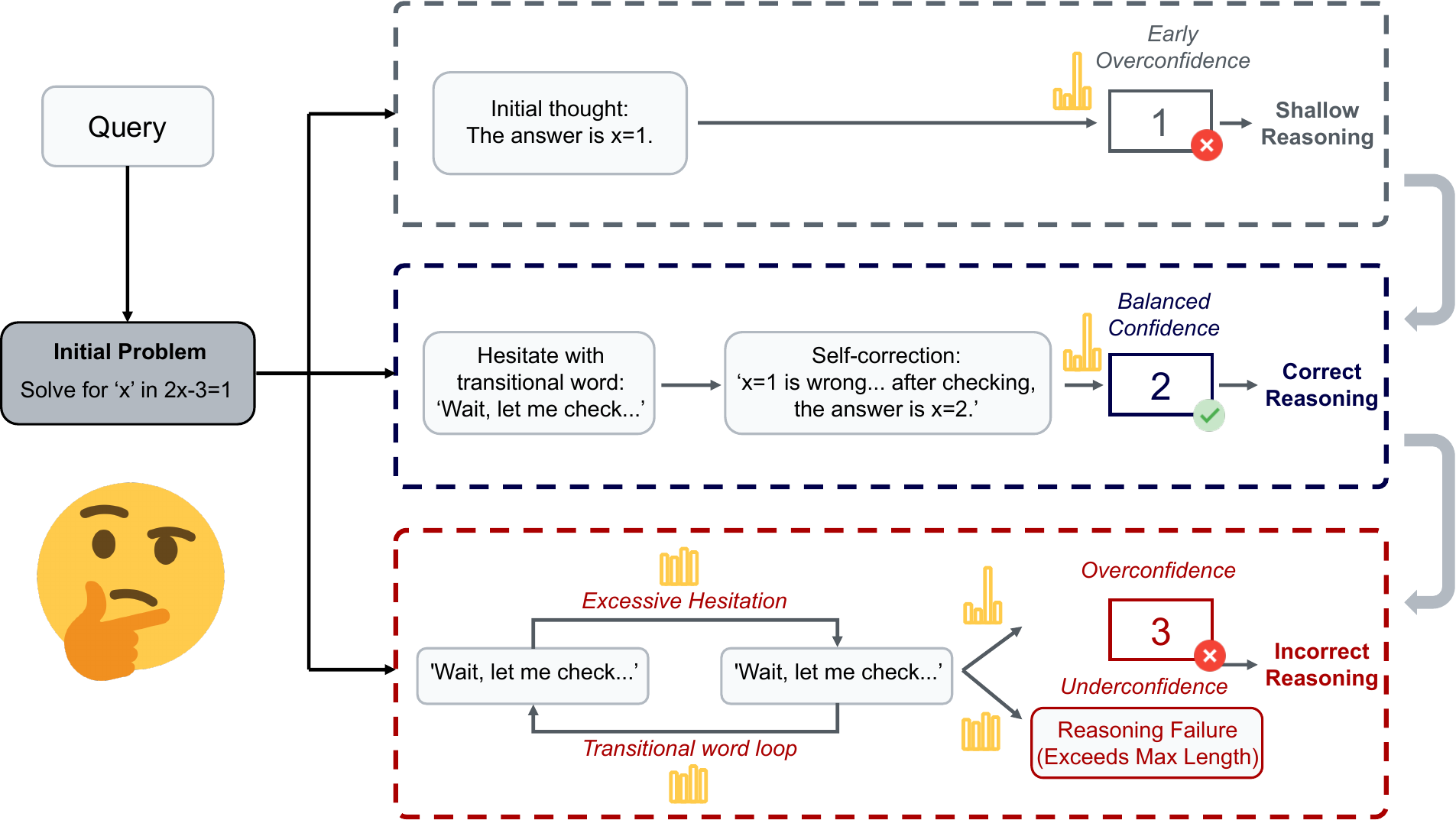}
    \caption{An example of the behavior of entropy and transitional words during the reasoning process. underconfidence and overconfidence both harm the performance of reasoning models.}
    \label{fig:pipeline}
\end{figure}

Our study is centered on a fundamental question: Under what conditions does the RLIF method work, and which internal feedback signal provides the best performance gain? In this work, we systematically explore the use of multiple forms of internal feedback—(1) self-certainty\citep{2025selfcertainty}, (2) trajectory-level entropy\citep{agarwal2025unreasonable}, and (3) token-level entropy\citep{agarwal2025unreasonable}—as replacements for ground-truth supervision in reinforcement learning. Through theoretical analysis, we establish the partial equivalence of these feedback signals under certain conditions, as they seem to optimize the same underlying objective, policy entropy. Then empirical results support this theory. Notably, our study demonstrates that while internal feedback signals significantly improve the performance of Base LLMs (such as Qwen2.5-3B, Qwen3-1.7B/4B, Qwen2.5-1.5B-Math), in contrast, instruct models—those that have undergone further human-aligned fine-tuning—tend to suffer performance degradation under RLIF training. We further propose and empirically validate an explanation for this performance gap. We attribute this divergence to differences in policy entropy: within the same model family, Base models generally exhibit higher policy entropy than their instruct counterparts, making them more adaptive to internal feedback optimization. To further validate this explanation, we employ model merging techniques to construct intermediary models between Base and instruct variants, thereby mapping the entropy and performance landscape. Additionally, our analysis reveals that the frequency of ``transitional words'' \citep{wang2025beyond} (logical connectors and transitional terms essential for multi-step reasoning) decreases during training.  This reduction appears to explain the phenomenon where RLIF-driven performance initially improves but then declines as training progresses. Overall, our findings not only delineate the precise conditions under which RLIF is beneficial, but also offer practical guidelines for its application in future LLM reinforcement learning setups.

\section{Related Works}
\label{sec:related_works}

\subsection{Reinforcement Learning for LLMs}
RL was first applied to LLMs through reinforcement learning with human feedback (RLHF), a post-training technique designed to align LLMs' outputs with human preferences and enhance their instruction-following capabilities. With the emergence of reasoning-focused LLMs \citep{openai2024reasoning,guo2025deepseek,team2025kimi}, RL has become indispensable for stimulating their reasoning abilities. DeepSeek-R1 \citep{guo2025deepseek}, one of the most advanced open-source reasoning LLMs, had its reasoning capabilities significantly enhanced through GRPO \citep{shao2024deepseekmath}, building upon its base model DeepSeek-V3 \citep{liu2024deepseek}. DeepSeek-R1 introduced the "zero RL" paradigm \citep{wang2025beyond}, which applies RL methods to directly stimulate reasoning abilities in LLMs from their pre-trained base models without supervised fine-tuning (SFT). This approach has been successfully adopted by QwQ \citep{qwq32b} and Qwen3 \citep{yang2025qwen3}. Compared to the previously dominant RL algorithm PPO \citep{schulman2017proximal}, GRPO eliminates the need for a critic model, making it remarkably cost-effective. Building on GRPO's success, several other critic-model-free RL algorithms have been proposed to enhance LLMs' reasoning capabilities, including DAPO \citep{yu2025dapo}, VAPO \citep{yue2025vapo}, and TRPA \citep{su2025trust}.

\subsection{Large Reasoning Model}
Large reasoning models (LRMs) aim to enhance multi-step reasoning in LLMs by generating and optimizing thought trajectories through prompting, reinforcement learning, and long-form inference. Building on Chain-of-Thought (CoT) \citep{wei2022chain} and self-reflection prompting \citep{liu2024self_reflection}, recent work explores training-time RL extensions and test-time long-form strategies \citep{xu2025reinforcedreasoning}. DeepSeek-R1 leveraged GRPO’s ``zero RL'' paradigm to significantly boost reasoning performance \citep{shao2024deepseekmath,wang2025beyond}, while AM-Thinking-v1 (32B) achieved state-of-the-art results on AIME 2024 and LiveCodeBench, rivaling much larger MoE models \citep{ji2025amthinking}. However, \citet{shojaee2025illusion}'s study on the ``illusion of thinking'' demonstrated that LRMs can suffer accuracy collapse on highly complex problems, underperform standard LLMs on low-complexity tasks, and offer only marginal gains on intermediate benchmarks. Similarly, evaluation on the 2025 USA Mathematical Olympiad showed proof success rates below 5\% \citep{petrov2025proof_or_bluff}. These findings expose fundamental scalability challenges and underscore the need for novel architectural and training paradigms.

\subsection{Reinforcement Learning from Internal Feedback}
Reinforcement Learning from Internal Feedback (RLIF) was first introduced, to the best of our knowledge, by \citet{zhang2025right}, who employed clustering techniques to extract semantic entropy across multiple rollouts and compute corresponding advantages. Building on this direction, \citet{2025selfcertainty} proposed leveraging unsupervised signals—derived directly from the model's own outputs—as reward signals for reinforcement learning updates in large language models (LLMs). Central to their framework is the concept of self-certainty, which quantifies the model's internal confidence in its generated predictions. Other studies have similarly explored unsupervised metrics that reflect a model’s output certainty, including token-level entropy \citep{agarwal2025unreasonable}, trajectory-level entropy \citep{agarwal2025unreasonable}, and self-confidence \citep{li2025confidenceneedfewshotrl}. These metrics collectively measure the concentration of the model’s output distribution—i.e., the extent to which the predicted probabilities approximate a one-hot distribution, indicating high certainty. In a related effort, TTRL \citep{zuo2025ttrl} interpreted majority-vote outcomes as labels for computing advantages. While it did not directly utilize entropy-based measures, its objective also centered on optimizing the model’s output distribution toward a more deterministic (i.e., one-hot) form. The use of such internal signals as reward functions offers a promising direction for reinforcement learning in LLMs, enabling fully unsupervised training paradigms that remove the dependency on external feedback.

\section{Preliminaries}

\subsection{Group Relative Policy Optimization (GRPO)}
GRPO~\citep{shao2024deepseekmath} is one of the most widespread RLVR algorithms, which extends PPO by removing the value network and using group-based rewards to estimate advantages. Instead of relying on a critic, GRPO computes relative advantages by comparing sampled responses within the same group. For each input $\boldsymbol{x}$ with its ground-truth answer $\boldsymbol{a}$, the rollout policy $\pi_{\boldsymbol{\theta}_{\text{old}}}$ generates a set of $G$ sampled responses $\{\boldsymbol{o}^i\}_{i=1}^{G}$. The estimated advantage $\hat{A}^i_t$ is then computed as:
\begin{equation}
\hat{A}^i_t = \frac{r^i - \text{mean}(\{R^i\}_{i=1}^{G})}{\text{std}(\{R^i\}_{i=1}^{G})}, \quad \text{where}\ \  R^i = \begin{cases}
1.0 & \quad \text{if }\ \  \texttt{is\ equivalent}(\boldsymbol{a}, \boldsymbol{y}^i), \\
0.0 & \quad \text{otherwise}.
\end{cases} 
\label{eq: grpo advantage}
\end{equation}

Beyond this new advantage estimation, GRPO also modifies the PPO objective by adding a KL-divergence penalty between the current policy and a reference model, incorporating prior knowledge into training.

\subsection{RLIF Algorithms}
Strictly speaking, RLIF~\citep{2025selfcertainty} methods are not reinforcement learning algorithms in the traditional sense, such as GRPO or PPO. Instead, they comprise a set of techniques developed in the context of reinforcement learning for LLMs, which convert external reward signals into internally generated, unsupervised feedback.

The primary approaches include self-certainty~\citep{2025selfcertainty}, token-level entropy~\citep{agarwal2025unreasonable}, trajectory-level entropy~\citep{agarwal2025unreasonable}, and majority voting. These methods provide alternative reward signals without relying on externally supervised feedback.

In this paper, we focus on three of these representative approaches—self-certainty, token-level entropy, and trajectory-level entropy—leaving majority voting for future investigation. The following sections will introduce each of these three reward design approaches in detail.

\begin{table}[ht]
\resizebox{\columnwidth}{!}{
\begin{tabular}{lll}

\hline
                         & Formula                                                                                                                & Papers                                                    \\ \hline
self-certainty           & $r_{\text{self-certainty}}(x,y) = \frac{1}{|y|}\sum_{t=1}^{|y|}\displaystyle \KL (U \Vert \pi_\theta(\cdot|x,y_{<t}))$ & \citep{2025selfcertainty}    \\
token-level entropy      & $r_{\text{token-entropy}}(x,y) = -\frac{1}{|y|}\sum_{t=1}^{|y|}\displaystyle H(\pi_\theta(\cdot|x,y_{<t}))$            & \citep{agarwal2025unreasonable}          \\
trajectory-level entropy & $r_{\text{traj-entropy}}(x,y) =\frac{1}{|y|}\log \pi_\theta(y|x)$                                                      & \citep{agarwal2025unreasonable}(revised) \\ \hline
\end{tabular}
}
\caption{Summary of RLIF Methods. This table presents the formulas and corresponding references for the three RLIF methods used in this paper.  Note that the formula for trajectory-level entropy has been modified from the original paper by the addition of the normalization factor $\frac{1}{|y|}$.}
\label{tab:my-table}

\end{table}

\begin{equation}
r_{\text{self-certainty}}(x,y) = \frac{1}{|y|}\sum_{t=1}^{|y|}\displaystyle \KL (U \Vert \pi_\theta(\cdot|x,y_{<t}))
\label{eq: self-certainty reward}
\end{equation}
\begin{equation}
r_{\text{token-entropy}}(x,y) = -\frac{1}{|y|}\sum_{t=1}^{|y|}\displaystyle H(\pi_\theta(\cdot|x,y_{<t}))
\label{eq: token-entropy reward}
\end{equation}
\begin{equation}
r_{\text{traj-entropy}}(x,y) = \frac{1}{|y|}\sum_{t=1}^{|y|}\log \pi_\theta(y_t|x,y_{<t})=\frac{1}{|y|}\log \pi_\theta(y|x)
\label{eq: traj-entropy reward}
\end{equation}

\subsection{Policy entropy}
Policy entropy measures the uncertainty in the actions selected by an agent.
Given a policy model $\pi_\theta$ and a training dataset $\mathcal{D}$, we define the average token-level entropy of the policy over the dataset as follows:
\begin{align}
\displaystyle H(\pi_\theta, \mathcal{D}) &= -\mathbb{E}_{\mathcal{D}, \pi_\theta} \left[ \log \pi_\theta(y_t | x,\boldsymbol{y}_{<t}) \right] \\
&= -\frac{1}{|\mathcal{D}|} \sum_{x \in \mathcal{D}} \frac{1}{|\boldsymbol{y}|} \sum_{t=1}^{|\boldsymbol{y}|} \mathbb{E}_{y_t \sim \pi_\theta} \left[ \log \pi_\theta(y_t | x, \boldsymbol{y}_{<t}) \right]\\
&=-\frac{1}{|\mathcal{D}|} \sum_{x \in \mathcal{D}} \frac{1}{|\boldsymbol{y}|} \sum_{t=1}^{|\boldsymbol{y}|}  \sum_{j\in V} \pi_\theta(j | x, \boldsymbol{y}_{<t}) \log \pi_\theta(j | x, \boldsymbol{y}_{<t}) 
\end{align}
Here, $V$ denotes the vocabulary, and $|\boldsymbol{y}|$ is the length of the output sequence.

The performance of the model increases as the policy entropy decreases during the training process in RLVR experiments. \citet{cui2025entropymechanismreinforcementlearning} explains the phenomenon, and an empirical formula was discovered: $\displaystyle R=-a\exp(\displaystyle H)+b$ , where $R$ means the performance of the model and $H$ indicates the policy entropy.

\subsection{What exactly does RLIF do?}
From both theoretical and experimental perspectives, it is straightforward to demonstrate that RLIF directly leads to a reduction in policy entropy, indicating that the model becomes increasingly ``confident'' and its output distribution approaches determinism. 

The theoretical mechanism behind the aforementioned phenomenon (referring to performance initially increasing then decreasing, and the decrease in entropy leading to reduced exploration) can be understood by analyzing the changes in policy entropy during the training process. Specifically, we elucidate how the RLIF method directly leads to a rapid drop in policy entropy through a series of lemmas and propositions.

To gain a deeper understanding of the dynamic changes in policy entropy within the RL process, particularly its relationship with parameter updates, we introduce the lemma \ref{lemma:ent_npg}. This lemma describes the impact of a single-step parameter update on the change in information entropy for a given state under a tabular softmax policy, revealing an approximate relationship between the change in entropy and the covariance of parameter updates:

\begin{lemma}
\label{lemma:ent_npg}
If $\pi_\theta$ is a tabular softmax policy, the difference of policy entropy between two consecutive steps satisfies
\[
\displaystyle{H}(\pi_\theta^{k+1}) - \displaystyle{H}(\pi_\theta^k) \approx - \mathbb{E}_{s\sim d_{\mu}^{\pi^{k}}}\mathrm{Cov}_{a\sim\pi^k_\theta(\cdot|s)}\left(\log\pi^k_\theta(a|s)~, \theta^{k+1}_{s,a} - \theta^k_{s,a}\right)
\]
\end{lemma}
This lemma was first introduced in~\citet{zhihu2025entropy}, and is summarized by~\citet{cui2025entropymechanismreinforcementlearning}. See proof in~\citet{zhihu2025entropy} and~\citet{cui2025entropymechanismreinforcementlearning}. Lemma \ref{lemma:ent_npg} provides a theoretical basis for us to analyze how the policy update affects entropy. Based on this, we can further discuss how different reward function settings in RLIF directly or indirectly impact policy entropy. As shown in the proposition below, several typical reward function designs used in RLIF essentially drive the model to minimize its policy entropy.

\begin{proposition}
\label{prop:token_ent}
Using $r_{\text{token-entropy}}(x,y) = -\frac{1}{|y|}\sum_{t=1}^{|y|}\displaystyle H(\pi_\theta(\cdot|x,y_{<t}))$ as the reward function in the LLM RL setting objective:$\arg\max_{\theta} \mathbb{E}_{x\sim D,y\sim \pi_\theta(\cdot|x)}[r(x,y)]$ is equivalent to minimizing the policy entropy. (see the proof in Appendix~\ref{appx:proof_token_ent})
\end{proposition}

\begin{proposition}
\label{prop:self_certainty}   
Assume that $\pi_\theta$ is a tabular softmax policy, then using $$r_{\text{self-certainty}}(x,y) = \frac{1}{|y|}\sum_{t=1}^{|y|}\displaystyle \KL (U \Vert \pi_\theta(\cdot|x,y_{<t}))$$ in RL leads to a reduction in policy entropy. (see the proof in Appendix~\ref{appx:proof_self_certainty})
\end{proposition}

\begin{proposition}
\label{prop:traj_ent}   
Assume that $\pi_\theta$ is a tabular softmax policy and the parameters are updated by Natural Policy Gradient~\citep{kakade2001}, then using $r_{\text{traj-entropy}}(x,y) = \frac{1}{|y|}\sum_{t=1}^{|y|}\log \pi_\theta(y_t|x,y_{<t})=\frac{1}{|y|}\log \pi_\theta(y|x)$ as a reward function in the LLM RL objective: $\arg\max_{\theta} \mathbb{E}_{x\sim D,y\sim \pi_\theta(\cdot|x)}[r(x,y)]$ leads to a reduction in the policy entropy. (see the proof in Appendix~\ref{appx:proof_traj_ent})
\end{proposition}

\section{Experiments}
\label{sec:experiments}
\begin{figure}[htbp]
    \centering
    
    \begin{subfigure}[b]{0.32\textwidth}
        \includegraphics[width=\linewidth]{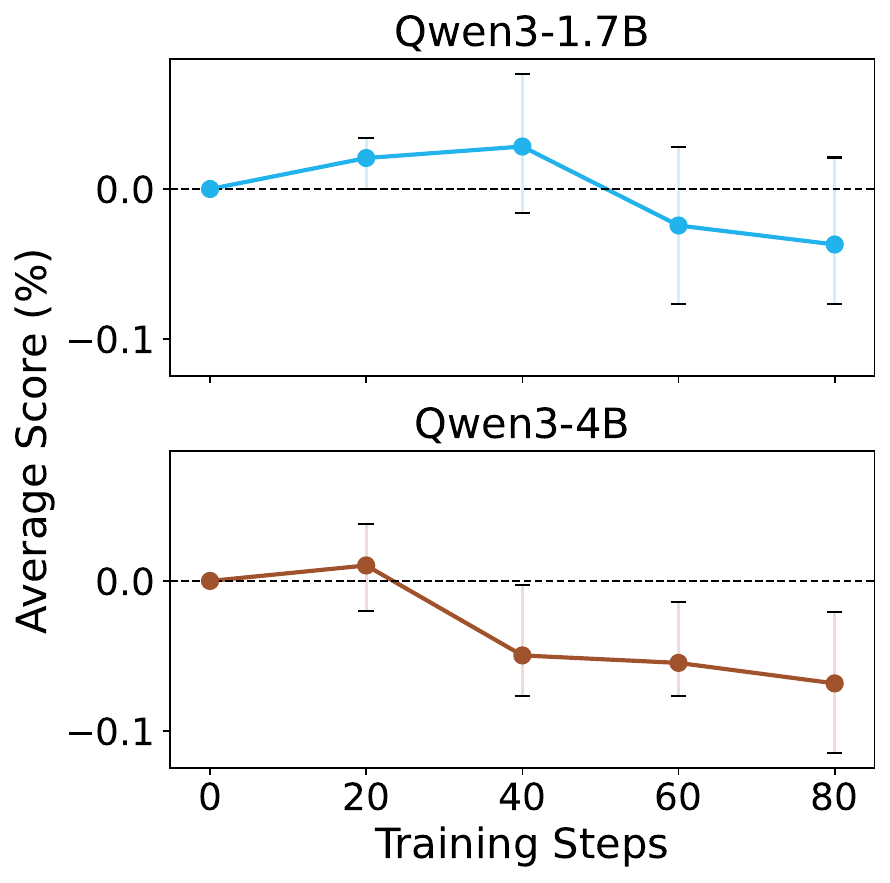}
        \caption{self-certainty}
        \label{fig:qwen25_3b_aime}
    \end{subfigure}
    \hfill
    \begin{subfigure}[b]{0.32\textwidth}
        \includegraphics[width=\linewidth]{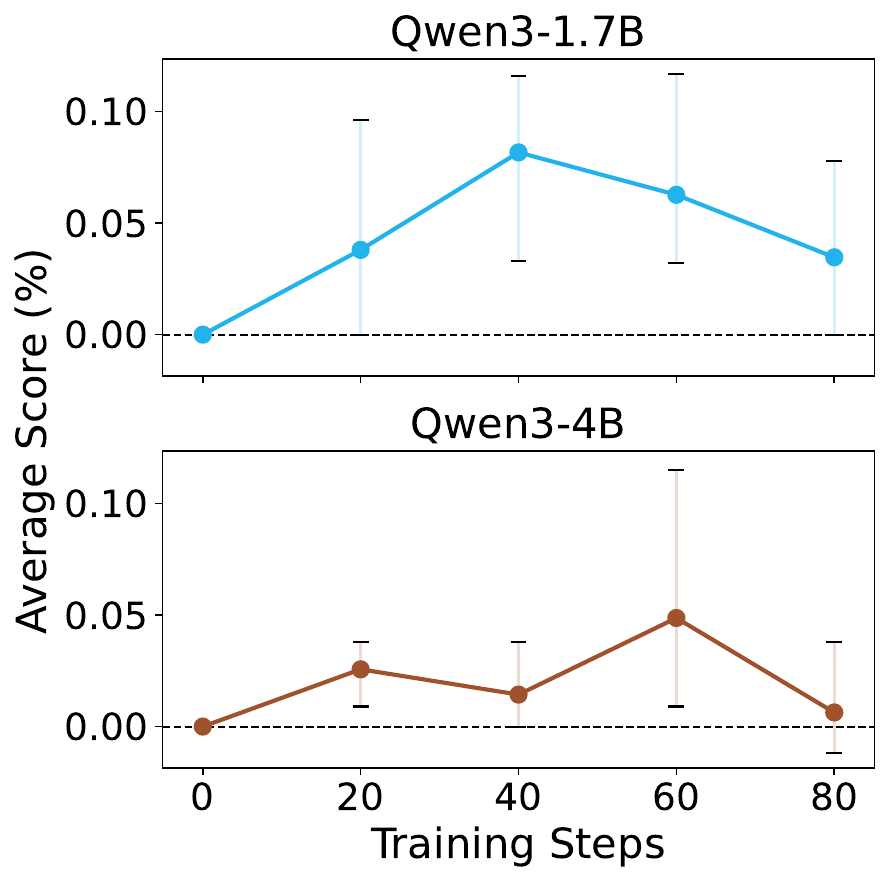}
        \caption{trajectory-level entropy}
        \label{fig:qwen25_3b_math}
    \end{subfigure}
    \hfill
    \begin{subfigure}[b]{0.32\textwidth}
        \includegraphics[width=\linewidth]{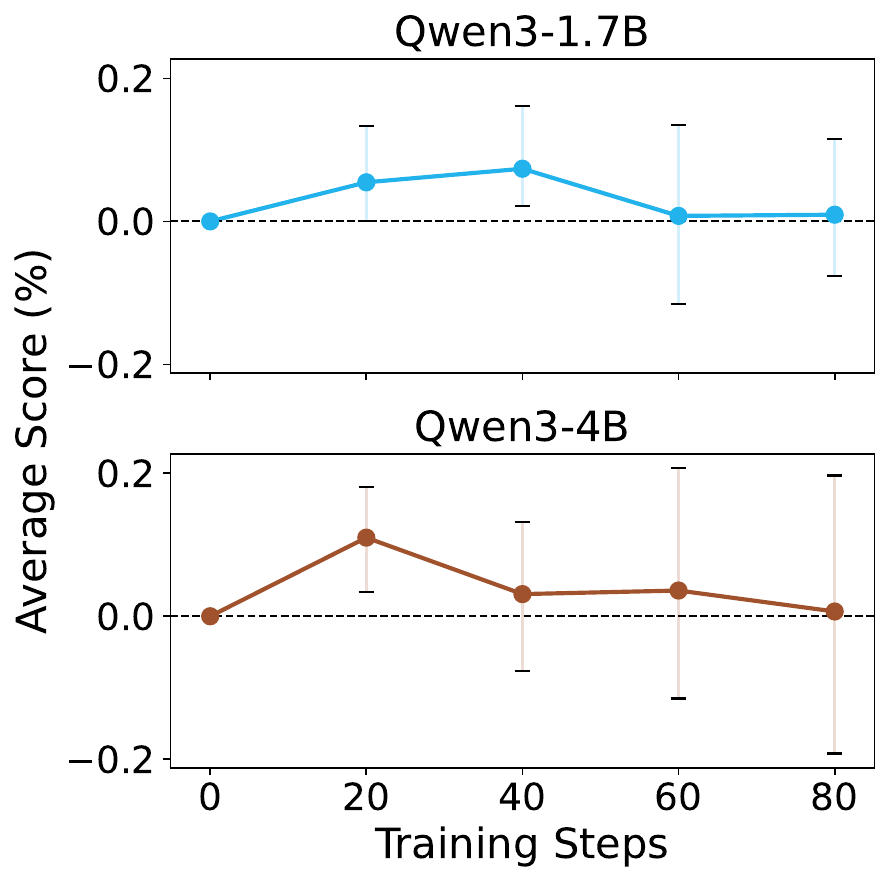}
        \caption{token-level entropy}
        \label{fig:qwen25_3b_gsm}
    \end{subfigure}

    \caption{Accuracy improvements relative to the base model on validation datasets (AIME2025, MATH500, and GSM8K) across multiple training steps. Different colors represent different base models (e.g., \textcolor{qwen3-4B}{Qwen3-4B}, \textcolor{qwen3-1.7B}{Qwen3-1.7B}). See Table~\ref{tab:main_chart} for the full numerical results.}
\label{fig:all_models}
\end{figure}
In this section, we conduct unsupervised RL experiments. First, we illustrate our experiment setup in Sec.\ref{sec:setup}. Second, we report the training and validation results in Sec.\ref{sec:results}. Last and most importantly, we give a detailed explanation of our results in Sec.\ref{sec:discussion}.
\subsection{Experiment Setup}
\label{sec:setup}
\paragraph{Benchmarks}
We evaluate all models on three widespread benchmarks. AIME2025 \citep{balunovic_srimatharena_2025} is extracted from the AIME2025 competition, which is an unseen dataset in the pretraining steps of all models mentioned in our work. The questions include geometry, number theory, algebra, and combinatorics. MATH500 \cite{balunovic_srimatharena_2025}  contains a subset of 500 problems from the MATH benchmark proposed by \citet{lightman2023let}. GSM8K \citep{cobbe2021gsm8k} is a dataset of 8.5K high-quality linguistically diverse grade school math word problems. 

\paragraph{Training}
All experiments are conducted on VeRL \citep{sheng2024hybridflow} framework on the training split of the MATH dataset \citep{hendrycks2021measuringmathematicalproblemsolving}, which contains 7,500 problems.
Each update processes $128$ problems, generating $8$ candidate solutions($G=8$ in GRPO) per problem, with a default KL penalty with reference model of $\beta = 0.005$. We respectively apply Qwen2.5-3B \citep{qwen2.5}, Qwen3-1.7B and Qwen3-4B \citep{yang2025qwen3} as our base model. We train our models using three RLIF methods: Self-certainty \citep{zhao2025learning}, Trajectory-level Entropy \citep{agarwal2025unreasonable}, and Token-level Entropy \citep{agarwal2025unreasonable}. For more details, please refer to Sec.\ref{sec:more_details}.

\subsection{The performance during RLIF training}
\label{sec:results}

By conducting three types of RLIF training on different models, we observed the performance of base models in Table \ref{tab:main_chart}: in the early stages of training (generally before 20 steps), model performance increases with the number of gradient steps. However, after this point, performance gradually declines, and in some method-model combinations, we observe instances of training collapse, characterized by extremely large gradient norms and a sudden, drastic decrease in response length. For Qwen2.5-Math, its performance on validation datasets continues to increase while training.

\subsection{Exploring the mechanism of behaviors during RLIF training.}
We first explore deeper into the performance increase in the early training stages, as shown in Table~\ref{tab:MATH500-answeranalysis}. From the results, we can summarize that the increase in the model's performance is mainly due to the enhancement of its instruction-following ability, while its reasoning ability increases a little. The model's reasoning capability decreases after 20 steps, which can be demonstrated by the percentage of WA (regardless of the correctness of the format).

We implement the idea of ``transitional words" \citep{wang2025beyond} to estimate the reasoning procedure of models. Transitional words are logical connectors and transitional terms—that are crucial for multi-step reasoning. For instance, ``but" and ``wait" indicate the transition against previous analysis, while ``since" and ``because" indicate more detailed reasoning of the previous proposal. \citet{wang2025beyond} proposed that transitional words often have high entropy during inference, and Table \ref{tab:fork} suggests that entropy minimization causes a reduction in the frequency of transitional words. The results will be discussed in more detail in the next section. 

\section{Discussion}
\label{sec:discussion}
\textbf{Claim 1: The effectiveness of those unsupervised reinforcement learning approaches vary across different models. Within the same model family, models with high initial policy entropy (such as base models) can be improved using the RLIF method. However, RLIF fails to improve models with low initial policy entropy.}

For most base models, the initial policy entropy tends to be relatively high, which corresponds to suboptimal instruction-following capabilities compared to instruct-tuned models. Consequently, RLIF methods can yield substantial improvements in performance, as demonstrated in Table~\ref{tab:main_chart}. As expected, we observe significant gains in the Qwen2.5-Math model, a finding corroborated by previous studies~\citep{gao2025oneshotentropyminimization, agarwal2025unreasonable}.

\begin{figure}[htbp]
    \centering
    \includegraphics[width=0.8\linewidth]{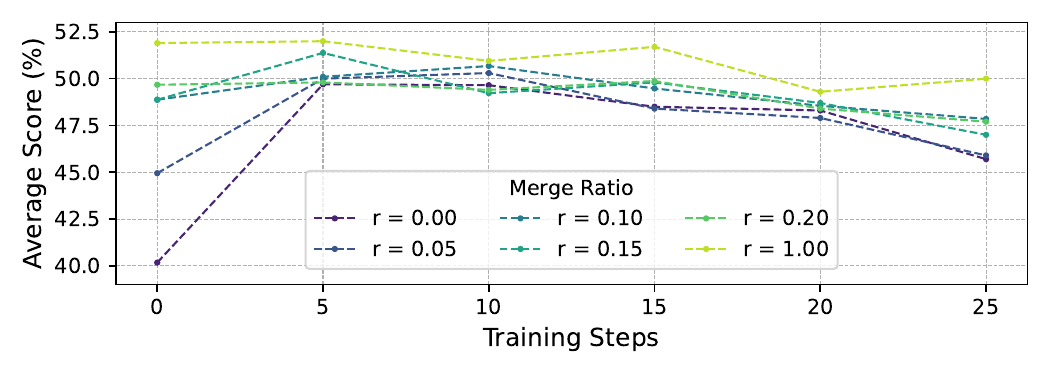}
    \caption{Averaged score (\%) of AIME2025, MATH500, and GSM8K during training. \(r\) denotes the merging ratio of the instruct model.}
    \label{fig:model_merging}
\end{figure}

To validate our hypothesis, we employ model merging, as introduced by \citet{yang2024modelmergingllmsmllms}, to construct a series of intermediate models interpolated between the base and instruct models. Specifically, we define a merge ratio as the weighting coefficient applied to the instruct model, denoted as $\text{merge\_ratio}=r \in [0, 1]$. The merged model parameters are then obtained via linear interpolation:
$$ \theta_\text{merged} = r \cdot \theta_\text{instruct} + (1 - r) \cdot \theta_\text{base} $$
These merged models are subsequently trained using self-certainty as the RLIF reward signal, and we visualize their performance evolution in Figure~\ref{fig:model_merging}. The model performance (average score) in the figure is the average accuracy on the test set of GSM8K, MATH500, Minervamath and OlympiadBench.
As shown in the figure, models with a smaller merge ratio (i.e., with a higher proportion of the base model) tend to have higher initial entropy and exhibit greater improvement during training. For instance, the base model (initial entropy 0.812), the model with merge ratio 0.1 (initial entropy 0.709), merge ratio 0.15 (initial entropy 0.489), and merge ratio 0.05 (initial entropy 0.828) all demonstrate significant gains. In contrast, models with a larger merge ratio — such as the instruct model (merge ratio 1.0, initial entropy 0.377) and the model with merge ratio 0.2 (initial entropy 0.436) — show little to no performance improvement during training.

\begin{table}[t]
\caption{Response Analysis of RLIF Evaluation on MATH500}
\label{tab:MATH500-answeranalysis}
\begin{tabularx}{\linewidth}{l l *{6}{>{\centering\arraybackslash}X}}
\toprule
 Model & steps & RA/WF & WA/WF & WA/RF & RA/RF & RA & WA \\
\hline
\multirow{3}{*}{Qwen2.5-3B}  & 0 & 31 & 36 & 169 & 260 & 291& 205\\
  & 20 & 1 & 2 & 204 & 288 & 289 & 206\\
  & 40 & 0 & 2 & 233 & 261 & 235& 261\\
\bottomrule
\end{tabularx}

\vspace{1mm}
{\footnotesize \textbf{Note:} RA = Right Answer, WA = Wrong Answer, RF = Right Format, WF = Wrong Format. \par}
\end{table}

\begin{table}[t]
    \centering
    \caption{The frequency of transitional words in the response of the validation set of MATH500 when training with the self-certainty method.}
    \begin{tabular}{c|ccccc}
    \toprule
         Model&  Step 0&  Step 20&  Step 40&  Step 60& Step 80\\
         \hline
         Qwen3-1.7B&  0.054&  0.044&  0.041&  0.038& 0.034\\
         Qwen3-4B &   0.053&  0.044&  0.042&  0.044& 0.044\\
         Qwen2.5-3B&  0.007&  0.004&  0.003&  0.002& 0.003\\
         Qwen2.5-3B-Instruct&  0.005&   0.003&    0.003&   0.002&    0.002\\   
    \bottomrule
    \end{tabular}
    \label{tab:fork}
\end{table}
\textbf{Claim 2: RLIF consistently reduces the occurrence of transitional words. Its performance improvements stem from mitigating underconfidence, whereas performance degradation arises from overconfidence.}

As shown in Table~\ref{tab:MATH500-answeranalysis}, the number of wrong-format (WF) responses decreases significantly, while the number of right-format (RF) responses increases as the number of RLIF steps grows. This trend suggests that performance gains observed from 0 to 20 RL steps primarily result from the model's improved ability to generate responses in the correct format. This improvement, illustrated by the example given in Sec.~\ref{appx:underover}, is due to the decreased response length and frequency of transitional words, so that an answer box can be generated and the number of ``WF" drops. However, Table~\ref{tab:MATH500-answeranalysis} suggests that the number of total right answers (regardless of the format) slightly decreases at the beginning of the training, and decreases significantly when the training continues (from 291 to 235). This suggests that the reasoning capability of the model is, in fact, degrading.

This phenomenon can be explained through the lens of entropy minimization. According to our proposition~\ref{prop:token_ent}, proposition~\ref{prop:self_certainty} and proposition~\ref{prop:traj_ent}, RLIF inherently reduces the policy entropy of the model, which leads to a substantial \textbf{decline in the generation of high-entropy tokens}. These high-entropy tokens often include frequently occurring transitional words. The suppression of such tokens reduces the model’s ability to explore alternate reasoning paths, thereby impairing its overall reasoning capability. Table~\ref{tab:fork} presents evidence from our analysis of transitional words on the Math500 dataset, demonstrating a clear downward trend in the occurrence of transitional words as training progresses.

For models that lack strong in-domain knowledge, unsupervised RL primarily functions to reduce the occurrence of transitional words and enhance the model’s inductive bias. In the early stages of training, this reduction helps to suppress underconfidence, allowing the model to better follow instructions and produce concise, goal-directed outputs within the maximum response length. However, as training continues and the transitional word frequency drops further, the model tends to become overconfident in its initial predictions, even when they are incorrect. This phenomenon—termed ``overconfidence"—is characterized by shallow reasoning and premature conclusion generation. As illustrated in Fig.~\ref{fig:pipeline}, both underconfidence and overconfidence impair the reasoning performance of LLMs. While entropy minimization via unsupervised RL successfully mitigates underconfidence, it inevitably shifts the model toward overconfidence.

By contrast, models with strong in-domain knowledge (e.g. finetuning with a massive amount of math data) benefit more robustly from RLIF. Their extensive prior knowledge and established reasoning skills enable them to maintain effective internal deliberation even as entropy decreases. Thus, while underconfidence is suppressed during training, these models do not exhibit the detrimental effects of overconfidence. Their performance remains stable or even improves, as they are better equipped to reason with fewer transitional words due to their stronger inherent knowledge base.
\section{Conclusion}

In this work, we present a unified theoretical framework that encompasses RLIF methods. We demonstrate that these methods are fundamentally partially-equivalent in nature. Furthermore, our empirical analysis reveals a consistent performance pattern across RLIF methods: performance initially improves but subsequently declines as training progresses. We attribute this phenomenon to a shift in the model’s reasoning behavior—specifically, from underconfidence to overconfidence. At the early stages of training, the model exhibits high entropy and tends to overanalyze, which leads to suboptimal generalization. As training continues, the model’s entropy decreases significantly, resulting in overly deterministic and shallow reasoning, i.e., overconfidence. This is demonstrated by our observation that the frequency of transitional words drops dramatically during training, resulting in the reduced thinking of the model. To further validate this hypothesis, we conduct experiments by interpolating between a base model (characterized by high entropy) and an instruct model (characterized by low entropy) using a range of mixing weights. The results confirm that the effectiveness of RLIF methods critically depends on the initial policy entropy level of the base model. Our findings highlight that the initial policy entropy is a key factor in the success of RLIF, and managing entropy appropriately is crucial to maintaining optimal model performance.

\bibliographystyle{abbrvnat}
\bibliography{references}

\appendix
\section{Appendix A}

\subsection{Useful Lemmas}


\begin{lemma}
\label{lemma:npg}
\textbf{Parameters update formula via Natural Policy Gradient for tabular softmax policy}
\begin{equation}
    \theta^{k+1}_{s,a}-\theta^{k}_{s,a}=\eta A^{\pi}(s,a)
\end{equation}
This lemma is from~\citet{JMLR:v22:19-736} and~\citet{zhihu2025entropy}.
\end{lemma}

\begin{lemma}
\textbf{Non-negative Covariance of a Probability Mass and its Logarithm}
\label{lemma:cov_p_logp_rigorous}

Let $\pi(\cdot\mid x)$ be a discrete probability distribution over a finite action space $\mathcal{Y}$. Let $y$ be a random variable drawn from this distribution, i.e., $y \sim \pi(\cdot\mid x)$. The covariance between the probability mass of an action and its logarithm is non-negative:
\[
\mathrm{Cov}_{y \sim \pi(\cdot\mid x)} \left[ \pi(y\mid x), \log \pi(y\mid x) \right] \ge 0
\]
Equality holds if and only if $\pi(y\mid x)$ is constant for all $y$ in the support of the distribution (i.e., $\pi$ is a uniform distribution over its support).
\end{lemma}

\subsection{Proof for Lemma~\ref{lemma:npg}} \label{appx:proof_npg}

We aim to show that, under the natural policy gradient (NPG) update rule, the parameter update at iteration \(k\) satisfies the following:

\begin{equation}
    \theta^{k+1}_{s,a} - \theta^{k}_{s,a} = \eta A^{\pi}(s,a),
\end{equation}

where \(\eta > 0\) is the step-size and \(A^{\pi}(s,a)\) is the advantage function under the current policy \(\pi\). This lemma is from~\citet{JMLR:v22:19-736} and~\citet{zhihu2025entropy}.

Consider the following objective at iteration \(k\). By the construction of the NPG algorithm, we aim to find a new policy \(\pi_{k+1}(\cdot\mid s)\) in the probability simplex \(\Delta(\mathcal{A})\) by maximizing
\[
  \max_{p(\cdot)\in\Delta(\mathcal{A})}
  \left\{
    \mathbb{E}_{a\sim p(\cdot)}\bigl[\,Q^{\pi^k}(s,a)\bigr]
    \;-\;
    \tfrac{1}{\eta}\,\mathrm{KL}\bigl(p(\cdot),\pi_k(\cdot\mid s)\bigr)
  \right\},
\]
where
\(\mathrm{KL}\bigl(p,\pi_k(\cdot\mid s)\bigr)=\sum_a p(a)\,\ln\frac{p(a)}{\pi_k(a\mid s)}\)
and \(\eta>0\) is a regularization parameter. Here, \(Q^{\pi^k}(s,a)\) is the action‐value function under the current policy \(\pi^k\).

We write \(p(a)\) for the new distribution over actions.  We must have \(p(a)\ge 0\) and \(\sum_a p(a)=1\).  Introduce a Lagrange multiplier \(\lambda\) for the normalization constraint, and define the Lagrangian
\[
  \mathcal{L}\bigl(p,\lambda\bigr)
  \;=\;
  \sum_{a} p(a)\,Q^{\pi^k}(s,a)
  \;-\;\frac{1}{\eta}\,\sum_a p(a)\,\ln\!\frac{p(a)}{\pi_k(a\mid s)}
  \;+\;\lambda\Bigl(\sum_{a} p(a)\;-\;1\Bigr).
\]
Our goal is to find \(p^*(a)=\pi_{k+1}(a\mid s)\) that maximizes \(\mathcal{L}\) subject to \(\sum_a p(a)=1\).

Differentiate \(\mathcal{L}\) with respect to \(p(a)\) and set to zero:
\[
  \frac{\partial \mathcal{L}}{\partial p(a)} 
  \;=\;
  Q^{\pi^k}(s,a)
  \;-\;\tfrac{1}{\eta}\,\ln\!\frac{p(a)}{\pi_k(a\mid s)}
  \;-\;\tfrac{1}{\eta}
  \;+\;\lambda
  \;=\;
  0.
\]
Rearranging,
\[
  \ln p(a)
  \;=\;
  \ln \pi_k(a\mid s)
  \;+\;
  \eta\,Q^{\pi^k}(s,a)
  \;-\;
  \eta\Bigl(\lambda\!+\!\tfrac{1}{\eta}\Bigr).
\]

Set \(\text{const} = \exp\{-\eta(\lambda + 1/\eta)\}\), which does not depend on \(a\).  Exponentiating yields
\[
  p(a)
  \;=\;
  \text{const}
  \;\times\;
  \pi_k(a\mid s)\,\exp\!\bigl\{\eta\,Q^{\pi^k}(s,a)\bigr\}.
\]

Finally, we enforce \(\sum_a p(a)=1\).  Hence
\[
  \text{const}
  \;=\;
  \biggl[
    \sum_{a'}\,\pi_k(a'\mid s)\,\exp\!\bigl\{\eta\,Q^{\pi^k}(s,a')\bigr\}
  \biggr]^{-1}.
\]

Thus
\[
  \pi_{k+1}(a\mid s)
  \;=\;
  \frac{\,\pi_k(a\mid s)\,\exp\!\bigl\{\eta\,Q^{\pi^k}(s,a)\bigr\}}
  {\,\sum_{a'}\;\pi_k(a'\mid s)\;\exp\!\bigl\{\eta\,Q^{\pi^k}(s,a')\bigr\}\!}.
\]

Observe that the advantage is often defined as \(A^{\pi^k}(s,a)=Q^{\pi^k}(s,a)-V^{\pi^k}(s)\).  Since \(V^{\pi^k}(s)\) is independent of \(a\), we can absorb \(\exp\{\eta\,V^{\pi^k}(s)\}\) into the normalizing denominator, yielding
\[
  \pi_{k+1}(a\mid s)
  \;\;\propto\;\;
  \pi_k(a\mid s)\,\exp\!\bigl\{\eta\,A^{\pi^k}(s,a)\bigr\}.
\]

Since we have:

\[
\pi_k(a|s) = \frac{\exp\left(\theta^k_{s,a}\right)}{\sum_{a' \in \mathcal{A}} \exp\left(\theta^k_{s,a'}\right)}.
\]

Substituting the softmax parameterization into this update:

\[
\pi_{k+1}(a|s) \propto \frac{\exp\left(\theta^k_{s,a}\right)}{\sum_{a' \in \mathcal{A}} \exp\left(\theta^k_{s,a'}\right)} \cdot \exp\left\{\eta A^\pi(s,a)\right\}.
\]

Simplifying the proportionality:

\[
\pi_{k+1}(a|s) \propto \exp\left(\theta^k_{s,a} + \eta A^\pi(s,a)\right).
\]

Normalizing to ensure \(\sum_{a} \pi_{k+1}(a|s) = 1\):

\[
\pi_{k+1}(a|s) = \frac{\exp\left(\theta^k_{s,a} + \eta A^\pi(s,a)\right)}{\sum_{a' \in \mathcal{A}} \exp\left(\theta^k_{s,a'} + \eta A^\pi(s,a')\right)}.
\]

So the parameter update rule is:

\[
\theta^{k+1}_{s,a} = \theta^k_{s,a} + \eta A^\pi(s,a).
\]

This completes the proof.

\subsection{Proof for Lemma~\ref{lemma:cov_p_logp_rigorous}}

The proof is based on the monotonicity of the logarithm function and the properties of expectation.

Let $y$ and $y'$ be two independent and identically distributed (i.i.d.) random variables, both drawn from the distribution $\pi(\cdot\mid x)$.
The logarithm function, $f(z) = \log z$, is monotonically increasing on its domain $(0, \infty)$. This implies that for any two positive numbers $z_1$ and $z_2$, the product of their differences $(z_1 - z_2)$ and $(\log z_1 - \log z_2)$ is nonnegative:\[
(z_1 - z_2)(\log z_1 - \log z_2) \ge 0
\]
Set $z_1 = \pi(y\mid x)$ and $z_2 = \pi(y^{\prime}\mid x)$. Since probabilities are nonnegative, we have:
\[
\left[ \pi(y\mid x) - \pi(y^{\prime}\mid x) \right] \left[ \log \pi(y\mid x) - \log \pi(y'\mid x) \right] \geq 0
\]
This inequality holds for any pair of outcomes $(y, y')$. We can now take the expectations of both sides with respect to the joint distribution of $y$ and $y'$. Since $y$ and $y'$ are i.i.d., the joint probability is $\pi(y\mid x)\pi(y'\mid x)$.
\[
\mathbb{E}_{y, y' \,\overset{\mathrm{i.i.d.}}{\sim}\, \pi(\cdot\mid x)}
\left[
  \left( \pi(y\mid x) - \pi(y'\mid x) \right)
  \left( \log \pi(y\mid x) - \log \pi(y'\mid x) \right)
\right] \ge 0
\]

Expanding the product inside the expectation gives:
\begin{align*}
\mathbb{E} \Biggl[ & \pi(y\mid x)\log\pi(y\mid x) - \pi(y\mid x)\log\pi(y'\mid x) \\
& - \pi(y'\mid x)\log\pi(y\mid x) + \pi(y'\mid x)\log\pi(y'\mid x) \Biggr] \ge 0
\end{align*}
By the linearity of expectation:
\begin{align*}
& \mathbb{E}[\pi(y\mid x)\log\pi(y\mid x)] - \mathbb{E}[\pi(y\mid x)\log\pi(y'\mid x)] \\
& - \mathbb{E}[\pi(y'\mid x)\log\pi(y\mid x)] + \mathbb{E}[\pi(y'\mid x)\log\pi(y'\mid x)] \ge 0
\end{align*}

Since $y$ and $y'$ are independent, the expectation of a product of functions of $y$ and $y'$ is the product of their expectations:
\[
\mathbb{E}[\pi(y\mid x)\log\pi(y'\mid x)] = \mathbb{E}[\pi(y\mid x)] \mathbb{E}[\log\pi(y'\mid x)]
\]
Furthermore, since $y$ and $y'$ are identically distributed. Then:
\begin{itemize}
    \item $\mathbb{E}[\pi(y\mid x)\log\pi(y\mid x)] = \mathbb{E}[\pi(y'\mid x)\log\pi(y'\mid x)]$
    \item $\mathbb{E}[\pi(y\mid x)] = \mathbb{E}[\pi(y'\mid x)]$
    \item $\mathbb{E}[\log\pi(y\mid x)] = \mathbb{E}[\log\pi(y'\mid x)]$
\end{itemize}
Substituting these into the inequality, we get:
\[
2 \mathbb{E}[\pi(y\mid x)\log\pi(y\mid x)] - 2 \mathbb{E}[\pi(y\mid x)] \mathbb{E}[\log\pi(y\mid x)] \ge 0
\]
Therefore, we have proven that:
\[
\mathrm{Cov}_{y \sim \pi(\cdot\mid x)} \left( \pi(y\mid x), \log \pi(y\mid x) \right) \ge 0
\]
The equality holds if and only if the inequality $(\pi(y|x) - \pi(y'\mid x))(\log \pi(y\mid x) - \log \pi(y'\mid x)) \ge 0$ is an equality for all pairs $(y, y')$ in the support. This happens only if $\pi(y\mid x) = \pi(y'\mid x)$ for all such pairs, which means the distribution is uniform over its support.

\subsection{Proof for proposition ~\ref{prop:token_ent}}\label{appx:proof_token_ent}
\begin{equation}
r_{\text{token-entropy}}(x,y) = -\frac{1}{|y|}\sum_{t=1}^{|y|}\displaystyle H(\pi_\theta(\cdot|x,y_{<t}))
\label{eq: token-entropy reward}
\end{equation}

It is straightforward to show that under the data distribution $D$ over $x$ and the model policy $\pi_\theta$ over $y$, the expected per-token entropy reward equals the expected entropy of the policy:
\[
\mathbb{E}_{x\sim D}\mathbb{E}_{y\sim\pi_\theta(\cdot|x)}[r_{\text{token-entropy}}(x,y)] = -H(\pi_\theta, D)
\]

is exactly the policy entropy over the data distribution.

Therefore, in the RL fine-tuning of LLMs, which can be considered a bandit RL setting, using $r_{\text{token-entropy}}$ as the reward will explicitly shape the policy update to minimize the average entropy of the model.

\subsection{Proof for Proposition~\ref{prop:self_certainty}}\label{appx:proof_self_certainty}

Assume that the policy $\pi_\theta$ is represented in a tabular softmax form. The Kullback–Leibler divergence between a uniform distribution $U$ over the action space $\mathcal{Y}$ and $\pi_\theta(\cdot\mid x)$ is given by $$
   \displaystyle \KL (U \,\|\, \pi_\theta(\cdot\mid x)) = \sum_{y\in\mathcal{Y}} \frac{1}{|\mathcal{Y}|} \log\frac{1/|\mathcal{Y}|}{\pi_\theta(y\mid x)}.
   $$
According to the tabular softmax structure, we have

\[
\begin{aligned}
    \frac{\partial \displaystyle    \displaystyle \KL (U\Vert \pi_\theta(\cdot\mid x)) }{\partial \theta_{x,y^{\prime}}} &=-\displaystyle  \sum_{y}\frac{1}{|\mathcal{Y}|}\frac{\partial \log\pi_\theta(y\mid x)}{\partial \theta_{x,y^{\prime}}} \\
    &=-\displaystyle  \sum_{y}\frac{1}{|\mathcal{Y}|} [\mathbf{1}_{\left\{y=y^{\prime}\right\}}-\pi_\theta\left(y^{\prime} \mid x\right)] \\
    &= -\displaystyle  \frac{1}{|\mathcal{Y}|}+\pi_\theta(y^{\prime}\mid x)
\end{aligned}
\]
Assuming a learning rate $\eta$, the update for the parameter $\theta_{x,y'}$ follows
\[
    \theta^{k+1}_{x,y^{\prime}}-\theta^{k}_{x,y^{\prime}}=\eta \left[-\frac{1}{|\mathcal{Y}|}+\pi_\theta(y^{\prime}\mid x)\right]
\]

Therefore,

\[
\begin{aligned}
\displaystyle{H}(\pi_\theta^{k+1}\mid x) - \displaystyle{H}(\pi_\theta^k\mid x) 
&\approx -\eta\cdot\mathrm{Cov}_{y\sim\pi^k_\theta(\cdot\mid x)}\left(\log\pi^k_\theta  (y\mid x)~,-\frac{1}{|\mathcal{Y}|}+\pi_\theta^{k}(y\mid x) \right)\\
&=-\eta\cdot\mathrm{Cov}_{y\sim\pi^k_\theta(\cdot\mid x)}\left(\log\pi^k_\theta(y\mid x)~,\pi_\theta^{k}(y\mid x) \right) \\
\end{aligned}
\]

Then we have,
\[\begin{aligned}
\displaystyle H\left(\pi_{\theta}^{k+1}\right)-\displaystyle H\left(\pi_{\theta}^{k}\right) & \approx \mathbb{E}_{x \sim d_{{\mu}}^{\pi^{k}}}\left[\displaystyle{H}\left(\pi_{\theta}^{k+1}| x\right)-\displaystyle{H}\left(\pi_{\theta}^{k}|x\right)\right] \\
&\approx -\eta \ \mathbb{E}_{x \sim d_{{\mu}}^{\pi^{k}}}\left[\operatorname{Cov}_{y \sim \pi_{\theta}^{k}(\cdot\mid x)}\left(\log \pi_{\theta}^{k}(y|x), \theta_{x, y}^{k+1}-\theta_{x, y}^{k}\right)\right] \\
&= -\eta \ \mathbb{E}_{x \sim d_{{\mu}}^{\pi^{k}}}\left[\operatorname{Cov}_{y \sim \pi_{\theta}^{k}(\cdot\mid x)}\left(\log \pi_{\theta}^{k}(y\mid x),\pi_{\theta}^{k}(y\mid x) \right)\right]\leq 0
\end{aligned}
\]

According to lemma~\ref{lemma:cov_p_logp_rigorous},
$\operatorname{Cov}_{y \sim \pi_{\theta}^{k}(\cdot\mid x)}\left(\log \pi_{\theta}^{k}(y\mid x),\pi_{\theta}^{k}(y\mid x) \right)\geq 0$.
So, if we use the self-certainty reward $r_{\text{self-certainty}}(x,y) = \frac{1}{|y|}\sum_{t=1}^{|y|}\displaystyle \KL (U \Vert \pi_\theta(\cdot\mid x,y_{<t}))$, the update process directly leads to reduction in policy entropy.

\subsection{Proof for Proposition~\ref{prop:traj_ent}}\label{appx:proof_traj_ent}
Assume that the policy parameters are updated via the Natural Policy Gradient method~\citep{kakade2001}. According to lemma~\ref{lemma:ent_npg} and lemma~\ref{lemma:npg}, under a first-order approximation, the change in the policy entropy at a given context $x$ can be expressed as
\[
\mathcal{H}(\pi_\theta^{k+1}\mid x) - \mathcal{H}(\pi_\theta^k\mid x) \approx -\eta\cdot\mathrm{Cov}_{y\sim\pi^k_\theta(\cdot\mid x)}\left(\log\pi^k_\theta(y\mid x)~, A(x,y)\right)
\]

\begin{equation}
r_{\text{traj-entropy}}(x,y) =\frac{1}{|y|}\log \pi_\theta(y\mid x)
\end{equation}

As shown above, there is an intrinsic connection between the covariance of the log-probability $\log\pi_\theta(y\mid x)$ and the advantage value $A(x, y)$: a strong positive correlation results in a decrease of policy entropy, while a negative correlation leads to an increase in entropy. 

In the bandit formulation of RL fine-tuning for large language models, where the reward $r(x,y)$ depends only on the current context $x$ and the generated sequence $y$, the advantage function $A(x,y)$ is essentially equivalent to the reward (up to a constant baseline shift). By adopting the reward function defined as
$$
   r_{\text{traj-entropy}}(x,y)=\frac{1}{|y|}\log\pi_\theta(y\mid x),
   $$
we induce a strong positive correlation between the reward and $\log\pi_\theta(y\mid x)$, thereby rendering the covariance term positive. Consequently, each update results in a decrease in the policy entropy.

\section{Appendix B}
\subsection{Implementation Details}
\label{sec:more_details}
For the GSM8K and MATH500 dataset, we set the maximum response length to 3072, and for the AIME2025 dataset, we set it to 20480. For pass@1 experiments, we apply greedy decoding, and for pass@N experiments, we set \(T=0.6\) and \(topp=0.95\) while decoding.
For all datasets, we follow the prompt format given in \citet{2025selfcertainty}, which is: 

\begin{tcolorbox}[colback=cyan!10, colframe=teal!80!black, title=User Prompt]
\textbf{[Question] }You are a helpful AI Assistant, designed to provide well-reasoned and detailed responses. You FIRST think about the reasoning process step by step and then provide the user with the answer. Please enclose your final answer in the box: \textbackslash \textbackslash boxed\{\}.
\end{tcolorbox}
\subsection{Full Results}

The full results of three RLIF methods on validation datasets are presented in Table~\ref{tab:main_chart}.
\begin{table}[]
\caption{Results of RLIF methods on vaious benckmarks during the training. "N/A" denotes training collapse, and "\#" denotes the reasoning process exceeds the model's maximum response length.}
\centering
\resizebox{\columnwidth}{!}{%
\label{tab:main_chart}
\begin{tabular}{cc|ccc|ccc|ccc}
\toprule
\multirow{2}{*}{Base Model} & \multirow{2}{*}{\begin{tabular}[c]{@{}c@{}}Training \\ Steps\end{tabular}} & \multicolumn{3}{c|}{AIME2025}      & \multicolumn{3}{c|}{MATH500}     & \multicolumn{3}{c}{GSM8K}       \\
                            &                                                                            & self-certainty& trajectory & token& self-certainty& trajectory & token& self-certainty& trajectory & token\\
                            \hline
\multirow{5}{*}{Qwen2.5-3B} & 0                                                                          & 0.0&            0.0&         0.0& 0.524    &            0.524    &         0.524    & 0.649    &            0.649    &         0.649    \\
                            & 20                                                                         & \textbf{0.038}&            N/A&         0.0& \textbf{0.583}&            N/A&         0.540& \textbf{0.778}&            N/A&         \textbf{0.771}\\
                            & 40                                                                         & 0.038    &            N/A&         N/A& 0.526    &            N/A&         N/A& 0.756    &            N/A&         N/A\\
                            & 60                                                                         & 0.038    &            N/A&         N/A& 0.496    &            N/A&         N/A& 0.701    &            N/A&         N/A\\
                            & 80                                                                         & 0.038    &            N/A&         N/A& 0.454    &            N/A&         N/A& 0.670    &            N/A&         N/A\\
                            \hline
\multirow{5}{*}{Qwen3-1.7B} & 0                                                                          & 0.269    &            0.269    &         0.269    & 0.587    &            0.587    &         0.587    &  0.836        &           0.836        &         0.836        \\
                            & 20                                                                         & 0.269    &            0.269&         0.269& \textbf{0.621}&            \textbf{0.683}&         0.720
&  \textbf{0.864}&            0.854&         \textbf{0.867}\\
                            & 40                                                                         & \textbf{0.346}&            \textbf{0.385}&         \textbf{0.308}& 0.571&            0.683&         \textbf{0.748}&  0.860&            \textbf{0.869}&         0.857\\
                            & 60                                                                         & 0.192    &            0.308&         0.154& 0.563         &            0.704&         0.722
& 0.864&            0.868&         0.839\\
                            & 80                                                                         & 0.192    &            0.269&         0.192& 0.532&            0.665&         0.702&  0.857&            0.862&         0.826\\
                            \hline
\multirow{5}{*}{Qwen3-4B}   & 0                                                                          & 0.500         &           0.500         &        0.500         & \textbf{0.591}&            0.591    &         0.591    & 0.883    &            0.883    &         0.883    \\
                            & 20                                                                         &  \textbf{0.538}&            0.538&         \textbf{0.615}& 0.571&            0.621&         0.722& \textbf{0.896}&            \textbf{0.892}&         0.917\\
                            & 40                                                                         &   0.423&            0.500&         0.423& 0.522&            \textbf{0.629}&         0.723& 0.880&            0.888&         \textbf{0.921}\\
                            & 60                                                                         & 0.423&            \textbf{0.615}&         0.385& 0.518&            0.613&         \textbf{0.798}& 0.869&            0.892&         0.899\\
                            & 80                                                                         & 0.385&            0.538&         0.308& 0.522&            0.579&         0.788& 0.862&            0.876&        0.898\\
\hline
\multirow{5}{*}{Qwen2.5-1.5B-Math}   & 0                                                                          & \#&           \#&        \#&     0.532&            0.532&         0.532& 0.265&            0.265&         0.265\\
                            & 20                                                                         &  \#&            \#&         \#& 0.630         &            0.619&         0.668& 0.742&            0.748&         0.762\\
                            & 40                                                                         &   \#&            \#&         \#& 0.646    &            0.637&         0.680& 0.772&            0.754&         0.786\\
                            & 60                                                                         & \#&            \#&         \#& 0.656    &            0.655&         0.700& 0.778    &            0.753&         0.813\\
                            & 80                                                                         & \#&            \#&         \#& \textbf{0.670}&            \textbf{0.665}&         \textbf{0.672}& \textbf{0.779}&            \textbf{0.766}&        \textbf{0.830}\\
\bottomrule
\end{tabular}%
}
\end{table}
\begin{table}[]
\caption{Results of RLIF methods on vaious benckmarks during the training. "N/A" denotes training collapse.}
\centering
\resizebox{\columnwidth}{!}{%
\label{tab:instruction_chart}

\begin{tabular}{cc|ccc|ccc|ccc}
\toprule
\multirow{2}{*}{Instruct Model} & \multirow{2}{*}{Training Steps} & \multicolumn{3}{c|}{AIME2025} & \multicolumn{3}{c|}{MATH500} & \multicolumn{3}{c}{GSM8K}  \\
                            &                                 & self-certainty & trajectory & token & self-certainty & trajectory & token & self-certainty & trajectory & token \\
\midrule
\multirow{5}{*}{Qwen2.5-3B-Instruct} 
    & 0    & 0.000 & \textbf{0.000}& \textbf{0.000}& \textbf{0.636}& \textbf{0.636}& \textbf{0.636}& \textbf{0.860}& \textbf{0.860}& \textbf{0.860}\\
    & 20   & 0.000 & 0.000 & 0.000& 0.621 & 0.579 & 0.619& 0.821 & 0.732 & 0.816\\
    & 40   & \textbf{0.038}& 0.000 & 0.000& 0.591 & 0.534 & 0.524& 0.802 & 0.766 & 0.773\\
    & 60   & 0.038 & 0.000 & N/A& 0.534 & 0.494 & N/A& 0.750 & 0.699 & N/A\\
    & 80   & 0.038 & N/A& N/A& 0.496 & N/A& N/A& 0.745 & N/A& N/A\\
\bottomrule
\end{tabular}

}
\end{table}

\subsection{Comparison Between Three RLIF Methods}
\begin{figure}[htbp]
    \centering
    \begin{subfigure}{0.32\textwidth}
        \centering
        \includegraphics[width=\linewidth]{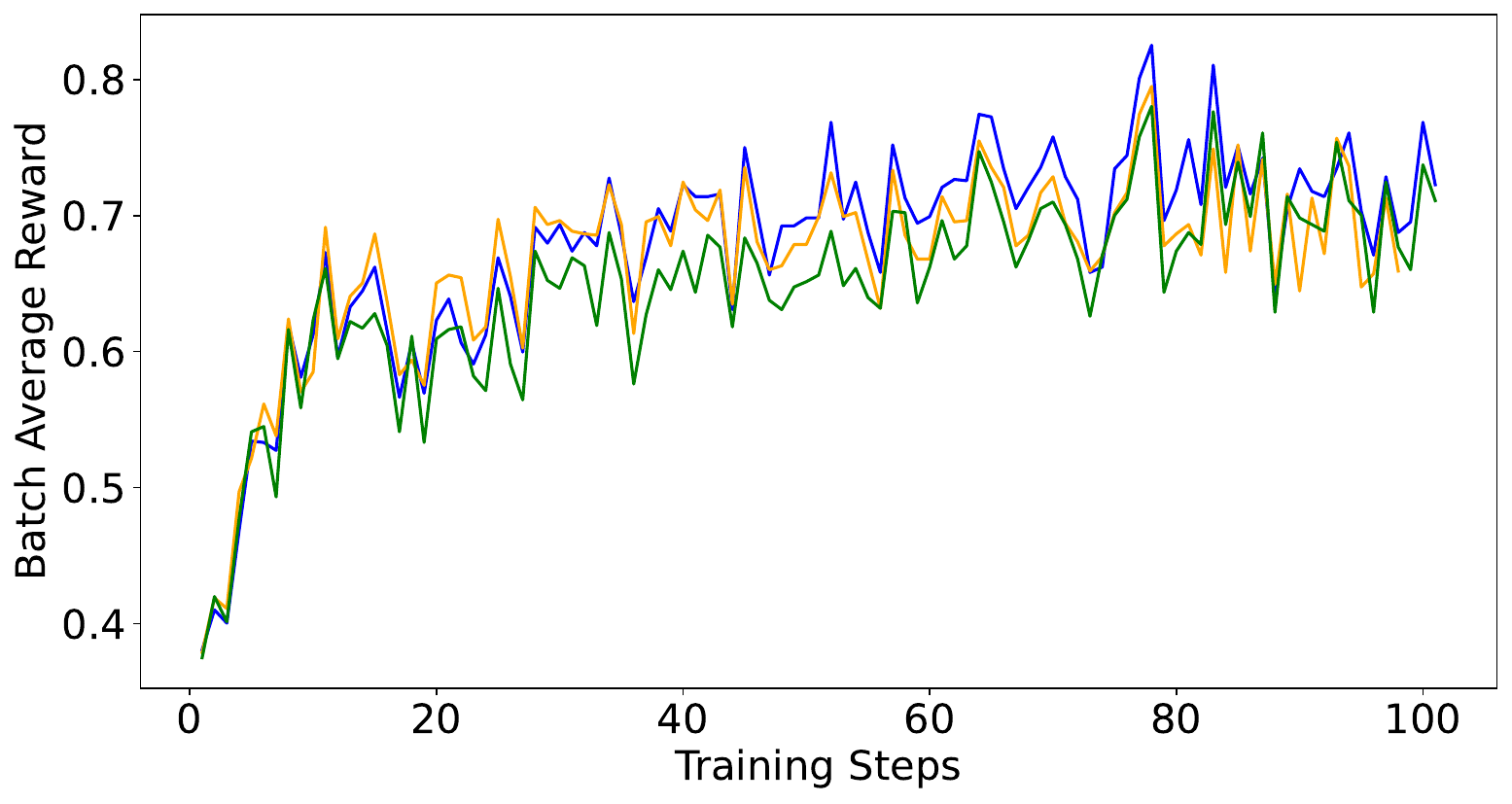}
        \caption{Batch Average Reward}
        \label{fig:critic_score_trends}
    \end{subfigure}
    \hfill
    \begin{subfigure}{0.32\textwidth}
        \centering
        \includegraphics[width=\linewidth]{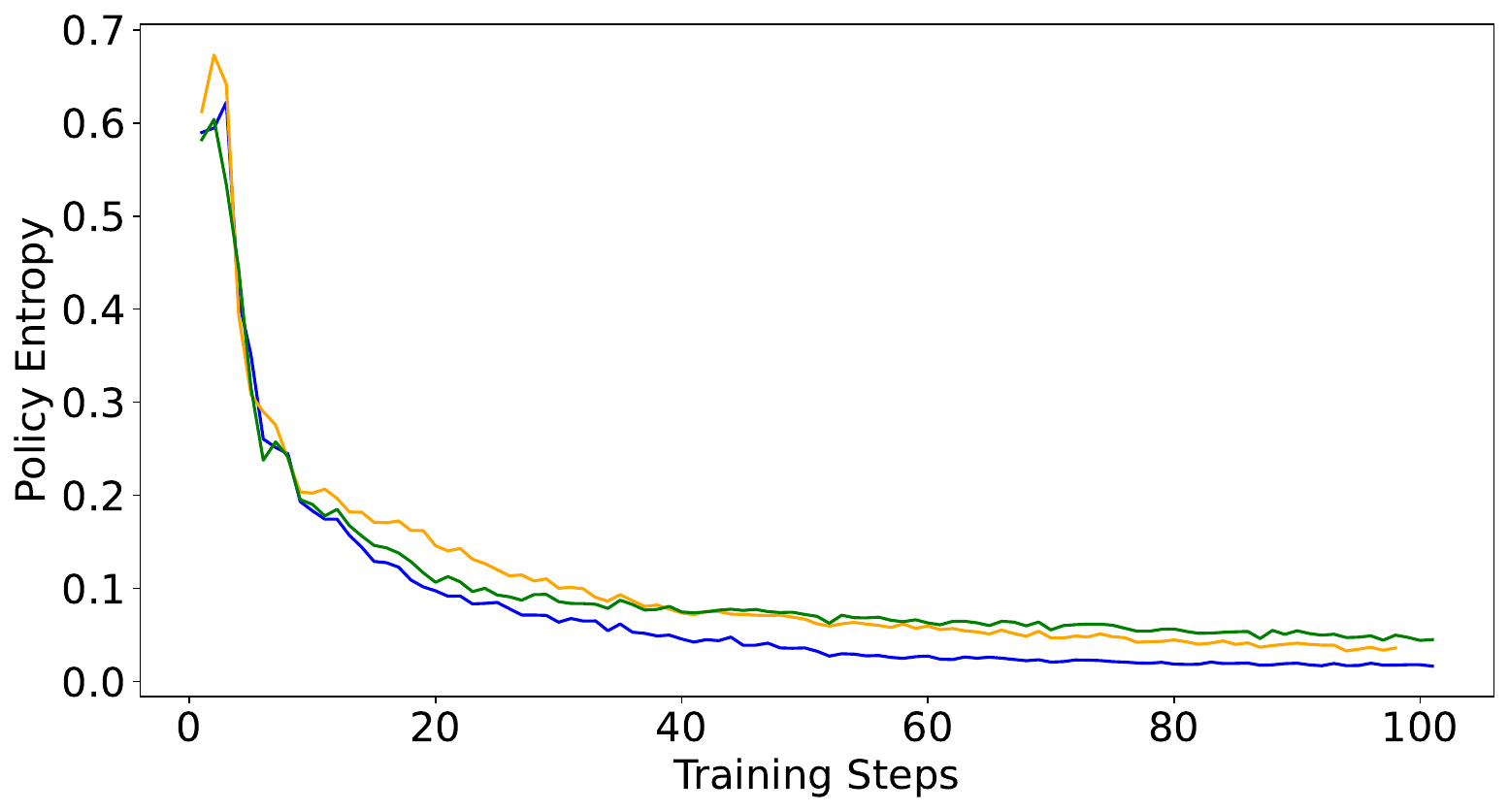}
        \caption{Policy Entropy}
        \label{fig:entropy_trends}
    \end{subfigure}
    \hfill
    \begin{subfigure}{0.32\textwidth}
        \centering
        \includegraphics[width=\linewidth]{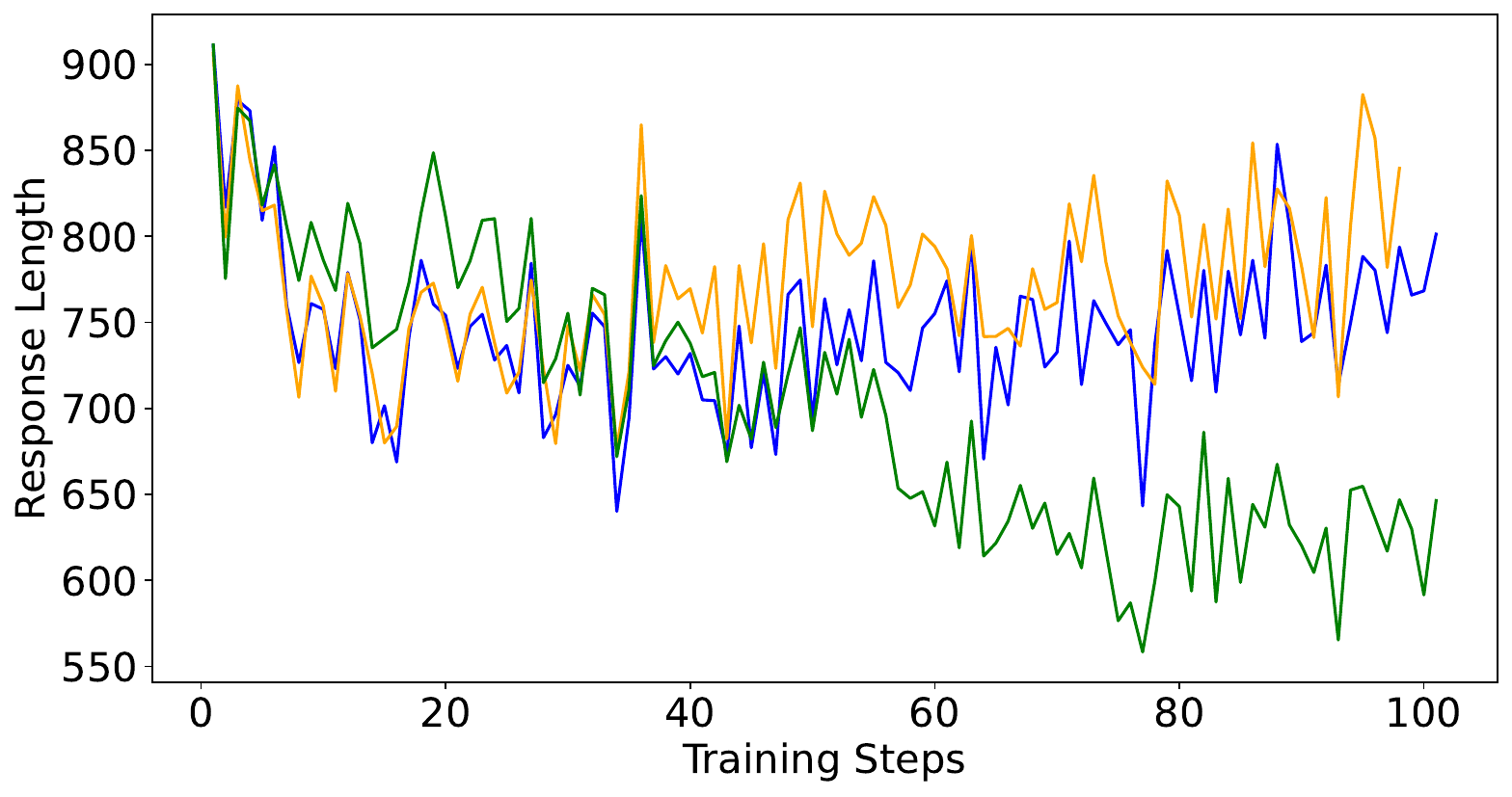}
        \caption{Response Length}
        \label{fig:response_length_trends}
    \end{subfigure}
    \caption{
        Training comparison of three methods.
        All subfigures share the same legend: 
        \textcolor{orange}{\textbf{Token-level Entropy}}, 
        \textcolor{blue}{\textbf{Trajectory-level Entropy}}, 
        \textcolor{green}{\textbf{Self-Certainty}}.
    }
    \label{fig:training_comparison}
\end{figure}
From the comparison of the above three figures, we observe that the policy entropy decreases steadily over the course of training. Moreover, the trends of policy entropy and the average reward on the training set are nearly identical. This phenomenon further indicates the ``equivalence'' among the three methods.
\subsection{Transitional Words}
By analyzing the entropy \citep{wang2025beyond} and semantic information of all output words, we give the following 14 words as transitional words. These words have high entropy, and they directly contribute to the logic transition in the reasoning process.
\begin{tcolorbox}[colback=cyan!10, colframe=teal!80!black, title=Transitional Words]
But, However, Wait, Suppose,
Assume, If, Let, Check, Unless, Perhaps, Maybe, Might, Alternatively, Recall
\end{tcolorbox}
\subsection{Examples of overconfidence and underconfidence}
\label{appx:underover}
\begin{tcolorbox}[colback=gray!5!white, colframe=gray!75!black, title=Question 1]
Find $\sin 20^\circ {}sin 40^\circ {}sin 60^\circ {}sin 80^\circ {}sin 100^\circ {}sin 120^\circ {}sin 140^\circ {}sin 160^\circ.$
\end{tcolorbox}

\begin{tcolorbox}[colback=red!5!white, colframe=red!75!black, breakable,title=An example of lack of confidence (Qwen3-1.7B)]

\textbf{Solution:} \textbf{[...After Long Reasoning Process of About 2900 Tokens]} Then the product is $(0.6736 *1.1133)/4\approx(0.6736 * 1.1133)\approx 0.749$. Then divided by 4 is $\approx 0.187$. \underline{But} earlier, we had $P = 3/16 \approx 0.1875$. So that's very close. So that seems to confirm that P is indeed $3/16$. Therefore, the original product is $(3/16)^2 = 9/256 \approx 0.03515625$ \textbf{[Which Is the Correct Answer]}.
\underline{But} \underline{let} me \underline{check} with another approach. \underline{Maybe} using complex numbers. \underline{Let} me \underline{recall} that the product of sines can be related to the imaginary parts of complex exponentials. \underline{But} that \underline{might} be more complicated. \underline{Alternatively}, use the formula for the product of multiple sines. \textbf{[Reach Max Response Length]}

\textbf{GroundTruth:} $\frac{9}{256}$
\end{tcolorbox}

\begin{tcolorbox}[colback=blue!5!white, colframe=blue!75!black, breakable,title=An example of balanced confidence(Qwen3-1.7B Token-level Entropy Step-40)]

\textbf{Solution:} To find the value of the product:
\[
\sin 20^\circ \sin 40^\circ \sin 60^\circ \sin 80^\circ \sin 100^\circ \sin 120^\circ \sin 140^\circ \sin 160^\circ,
\]
we use the identity:
\[
\sin(180^\circ - x) = \sin x.
\]
Applying this identity to the relevant terms, we have:
\[
\sin 100^\circ = \sin 80^\circ,\quad \sin 120^\circ = \sin 60^\circ,\quad \sin 140^\circ = \sin 40^\circ,\quad \sin 160^\circ = \sin 20^\circ.
\]
Substituting these into the original product gives:
\[
\sin 20^\circ \sin 40^\circ \sin 60^\circ \sin 80^\circ \sin 80^\circ \sin 60^\circ \sin 40^\circ \sin 20^\circ.
\]
Grouping like terms, this simplifies to:
\[
(\sin 20^\circ)^2 (\sin 40^\circ)^2 (\sin 60^\circ)^2 (\sin 80^\circ)^2,
\]
or
\[
\left( \sin 20^\circ \sin 40^\circ \sin 60^\circ \sin 80^\circ \right)^2.
\]

Next, we use the identity:
\[
\sin 20^\circ \sin 40^\circ \sin 80^\circ = \frac{\sin 60^\circ}{4}.
\]

Therefore,
\[
\sin 20^\circ \sin 40^\circ \sin 60^\circ \sin 80^\circ = \sin 60^\circ \times \frac{\sin 60^\circ}{4} = \frac{(\sin 60^\circ)^2}{4}.
\]
\underline{Recall} that
\[
\sin 60^\circ = \frac{\sqrt{3}}{2},
\]
so
\[
\left( \sin 20^\circ \sin 40^\circ \sin 60^\circ \sin 80^\circ \right)^2 = \left( \frac{(\sqrt{3}/2)^2}{4} \right)^2 = \left( \frac{3}{4 \times 4} \right)^2 = \left( \frac{3}{16} \right)^2 = \frac{9}{256}.
\]
So the value of the original product is

\textbackslash boxed\{\textbackslash dfrac\{9\}\{256\}\}

\textbf{GroundTruth:} $\frac{9}{256}$
\end{tcolorbox}

\begin{tcolorbox}[colback=gray!5!white, colframe=gray!75!black, title=Question 2]
\textbf{Problem:} Find the remainder when
\[
f(x) = x^{10} + 5x^9 - 8x^8 + 7x^7 - x^6 - 12x^5 + 4x^4 - 8x^3 + 12x^2 - 5x - 5
\]
is divided by $x^2 - 1$.
\end{tcolorbox}
\begin{tcolorbox}[colback=red!5!white, colframe=red!75!black, title=An example of lack of confidence (Qwen3-1.7B)]

\textbf{Solution:}\textbf{[...After Long Reasoning Process of About 2800 Tokens]} So yes, the remainder is $-13x + 3$. Therefore, the answer is \textcolor{red}{$-13x + 3$}.\textbf{[Which Is the Correct Answer]} \underline{But} \underline{maybe} the problem expects the answer in a different form? For example, sometimes people prefer to write it as $3 -13x$, \underline{but} both are equivalent. \underline{However}, since the question says "the remainder", and the remainder is a polynomial, so both forms are acceptable. \underline{But} \underline{maybe} the problem expects it in a certain way. \underline{However}, in the context of the question, since they didn't specify, either is okay. \underline{But} let me \underline{check} if there's a possible mistake in the sign.
\underline{Wait}, when I did the substitution, I had $R(x) = -13x + 3$. \underline{Let} me check with $x = 1$: $R(1) = -13(1) + 3 = -10$, which is correct. And $R(-1) = -13(-1) + 3 = 13 + 3 = 16$, which is correct. So that's correct.
\underline{Alternatively}, if I write the remainder as $3 -13x$, that's the same thing. \underline{But} \underline{maybe} the problem expects the answer in a specific format. \underline{However}, since the question says ``the remainder", and the remainder is a polynomial, both forms are correct. \underline{But} \underline{maybe} the answer is expected to be written as $-13x + 3$. \underline{However}, sometimes the leading coefficient is positive
\textbf{[Reach Max Response Length]}

\textbf{GroundTruth:} $-13x + 3$
\end{tcolorbox}

\begin{tcolorbox}[colback=blue!5!white, colframe=blue!75!black,breakable,title=An example of balanced confidence (Qwen3-1.7B Token-level Entropy Step-40)]

\textbf{Solution:}
Since the divisor $x^2 - 1$ is quadratic, the remainder is a linear polynomial: $R(x) = ax + b$.
We have $f(x) = (x^2 - 1)Q(x) + ax + b$, so:
$$
f(1) = a\cdot 1 + b = a + b, \qquad f(-1) = a\cdot(-1) + b = -a + b
$$
\underline{Let} us compute $f(1)$ and $f(-1)$:

1. Calculate $f(1)$:
\begin{align*}
f(1) &= 1^{10} + 5\cdot1^9 - 8\cdot1^8 + 7\cdot1^7 - 1^6 - 12\cdot1^5 + 4\cdot1^4 \\
&\quad - 8\cdot1^3 + 12\cdot1^2 - 5\cdot1 - 5 \\
&= 1 + 5 - 8 + 7 - 1 - 12 + 4 - 8 + 12 - 5 - 5 \\
&= -10
\end{align*}
So $a + b = -10$.

2. Calculate $f(-1)$:
\begin{align*}
f(-1) &= (-1)^{10} + 5\cdot(-1)^9 - 8\cdot(-1)^8 + 7\cdot(-1)^7 - (-1)^6 \\
&\quad - 12\cdot(-1)^5 + 4\cdot(-1)^4 - 8\cdot(-1)^3 + 12\cdot(-1)^2 \\
&\quad - 5\cdot(-1) - 5 \\
&= 1 - 5 - 8 - 7 - 1 + 12 + 4 + 8 + 12 + 5 - 5 \\
&= 16
\end{align*}
So $-a + b = 16$.

3. Solve for $a$ and $b$:
$$
\begin{cases}
a + b = -10 \\
-a + b = 16
\end{cases}
$$
Add the two equations:
$$
(a + b) + (-a + b) = -10 + 16 \implies 2b = 6 \implies b = 3
$$
Subtract one from the other:
$$
(a + b) - (-a + b) = -10 - 16 \implies 2a = -26 \implies a = -13
$$
\vspace{0.5em}
Final Answer:

\textbackslash boxed\{-13x + 3\}

\textbf{GroundTruth:} $-13x + 3$
\end{tcolorbox}

\begin{tcolorbox}[colback=orange!5!white, colframe=orange!75!black,breakable,title=An example of overconfidence (Qwen3-1.7B Token-Entropy Step-80)]

\textbf{Solution:}
To find the remainder when
$f(x) = x^{10} + 5x^9 - 8x^8 + 7x^7 - x^6 - 12x^5 + 4x^4 - 8x^3 + 12x^2 - 5x - 5$
is divided by $x^2 - 1$, we use the fact that $x^2 \equiv 1 \mod (x^2 - 1)$.
This allows us to reduce higher powers of $x$:
\begin{align*}
x^2 &\equiv 1, \\
x^3 &\equiv x \cdot x^2 \equiv x \cdot 1 = x, \\
x^4 &\equiv x^2 \cdot x^2 \equiv 1 \cdot 1 = 1, \\
x^5 &\equiv x^4 \cdot x \equiv 1 \cdot x = x, \\
x^6 &\equiv x^5 \cdot x \equiv x \cdot x = x^2 \equiv 1, \\
x^7 &\equiv x^6 \cdot x \equiv 1 \cdot x = x, \\
x^8 &\equiv x^7 \cdot x \equiv x \cdot x = x^2 \equiv 1, \\
x^9 &\equiv x^8 \cdot x \equiv 1 \cdot x = x, \\
x^{10} &\equiv x^9 \cdot x \equiv x \cdot x = x^2 \equiv 1.
\end{align*}
Substituting these into $f(x)$:
\begin{align*}
f(x) &\equiv x^{10} + 5x^9 - 8x^8 + 7x^7 - x^6 - 12x^5 + 4x^4 - 8x^3 + 12x^2 - 5x - 5 \\
&\equiv 1 + 5x - 8(1) + 7x - 1(1) - 12x + 4(1) - 8x + 12(1) - 5x - 5 \\
&= 1 + 5x - 8 + 7x - 1 - 12x + 4 - 8x + 12 - 5x - 5.
\end{align*}
Combining like terms:

Constants: $1 - 8 - 1 + 4 + 12 - 5 = 3$
Terms with $x$: $5x + 7x - 12x - 8x - 5x = -5x$

Thus, the remainder when $f(x)$ is divided by $x^2 - 1$ is:

\textbackslash boxed\{-5x + 3\}

\textbf{GroundTruth:} $-13x + 3$
\end{tcolorbox}

\begin{tcolorbox}[colback=gray!5!white, colframe=gray!75!black,breakable,title=Question 3]
\textbf{Problem:} 
A regular pentagon is rotated counterclockwise about its center. What is the minimum number of degrees it must be rotated until it coincides with its original position?
\end{tcolorbox}

\begin{tcolorbox}[colback=red!5!white, colframe=red!75!black, title=An example of lack of confidence (Qwen3-1.7B)]

\textbf{Solution:}
\textbf{[...After Long Reasoning Process of About 2600 Tokens]}  
\underline{Wait}, \underline{maybe} the confusion is between the rotation angle and the position. \underline{If} you rotate the pentagon by 72 degrees, it's not in the original position, \underline{but} \underline{if} you rotate it by 360 degrees, it is. So, the minimal angle is 360 degrees. \underline{But} that seems like the answer. \underline{However}, I think there's a misunderstanding here.

\underline{Alternatively}, \underline{maybe} the question is asking for the angle between the original position and the rotated position, \underline{but} that would be 72 degrees. \underline{But} the question says "until it coincides with its original position," so the angle must be such that after rotating by that angle, it's back to the original. So, the angle is 360 degrees. \underline{But} that seems like the answer.

\underline{But} I think I need to check with the concept of rotational symmetry. A regular pentagon has rotational symmetry of order 5, meaning that it can be rotated by 72 degrees to map onto itself. So, the minimal angle is 72 degrees. \underline{But} in that case, rotating by 72 degrees would map the pentagon onto itself, \underline{but} not necessarily to the original position. \underline{Wait}, \underline{but} if you rotate the pentagon by 72 degrees, it's a different position, \underline{but} it's still the same pentagon. So, does that mean it coincides
\textbf{[Reach Max Response Length]}

\textbf{GroundTruth:} $72$
\end{tcolorbox}

\begin{tcolorbox}[colback=blue!5!white, colframe=blue!75!black, breakable,title=An example of balanced confidence (Qwen3-1.7B Token-level Entropy Step-40)]

\textbf{Solution:}

Okay, so I need to figure out the minimum number of degrees a regular pentagon must be rotated counterclockwise about its center to coincide with its original position. \underline{Let} me think. 

First, a regular pentagon has five sides, right? So each central angle between two adjacent vertices should be 360 degrees divided by 5. \underline{Let} me calculate that. 360 divided by 5 is 72. So each vertex is 72 degrees apart around the center. 

\underline{If} you rotate the pentagon by 72 degrees, each vertex would move to the position of the next vertex. So after one rotation of 72 degrees, the pentagon would look exactly like it did before. Therefore, the minimum number of degrees needed is 72. 

\underline{Wait}, \underline{but} let me verify. Let's imagine the pentagon labeled vertices A, B, C, D, E in order. \underline{If} I rotate it 72 degrees counterclockwise, point A would move to where B was, B to where C was, and so on. So the shape would be the same. Therefore, yes, 72 degrees. 

\underline{But} \underline{maybe} I should check with a different approach. The rotational symmetry of a regular pentagon is five-fold, meaning it has five lines of symmetry and five rotational symmetries. The order of rotation is the number of degrees needed to rotate it back to its original position, which should be 360 divided by the number of sides. Since it's a pentagon, 360/5 = 72. 

Yes, that makes sense. So the answer should be 72 degrees.

\textbackslash boxed\{72\}

\textbf{GroundTruth:} $72$
\end{tcolorbox}

\end{document}